\documentclass[a4paper]{article}

\usepackage{INTERSPEECH2022}
\usepackage{xcolor}

\usepackage{url}
\usepackage{latexsym}
\usepackage{tabularx}
\usepackage{arydshln}
\usepackage{multirow}
\usepackage{rotating}
\usepackage{array,multirow,graphicx}
\usepackage{enumitem}
\usepackage{soulpos}
\usepackage{tikz}
\usepackage{pgfplots}
\usepackage{latexsym}
\usepackage{tabularx}
\usepackage{arydshln}
\usepackage{booktabs}
\usepackage{rotating}
\usepackage{threeparttable}
\usepackage[english]{babel}
\usepackage{blindtext}
\usepackage[shortcuts]{extdash}
\usepackage[normalem]{ulem}
\usepackage{tikz}
\usepackage{pgfplots}
\usepackage{hhline}

\title{\textit{M-Adapter}: Modality Adaptation for End-to-End Speech-to-Text Translation}

\name{Jinming Zhao, Hao Yang, Ehsan Shareghi, Gholamreza Haffari}
\address{
  Department of Data Science \& AI, Monash University}
\email{{first.last}@{monash.edu}}

\begin{document}

\maketitle
%


\begin{abstract}

End-to-end speech-to-text translation models are often initialized with pre-trained speech encoder and pre-trained text decoder. This leads to a significant training gap between pre-training and fine-tuning, largely due to the modality differences between speech outputs from the encoder and text inputs to the decoder. In this work, we aim to bridge the modality gap between speech and text to improve translation quality. We propose M-Adapter, a novel Transformer-based module, to adapt speech representations to text. While shrinking the speech sequence, M-Adapter produces features desired for speech-to-text translation via modelling global and local dependencies of a speech sequence. Our experimental results show that our model outperforms a strong baseline by up to 1 BLEU score on the Must-C En$\rightarrow$DE dataset.\footnote{Our code is available at https://github.com/mingzi151/w2v2-st.}

\end{abstract}
\noindent\textbf{Index Terms}: speech translation, modality adaptation
\section{Introduction}
\label{sec:introduction}
Speech-to-text translation (ST) is the task of translating audio signals in one language
to text in a foreign language. It has been conventionally approached with
a cascaded architecture comprising automatic speech recognition (ASR) and machine
translation (MT) components. The more recent end-to-end (E2E) approach has attracted great attentions, which involves an audio encoder taking audio signals as input and a text decoder producing a translated text.
The E2E approach alleviates the issues of error propagation and high latency with
in the cascaded methods
\cite{sperber2020speech, chen2021specrec}. That said, the E2E approach often requires large amounts of paired audios and translated text, which is not available
for most language pairs. A common practice to remedy the data scarcity issue is to pre-train the audio encoder and text decoder, and initialize the ST model with the pre-trained encoder and decoder~\cite{weiss2017sequence,tang2021improving}. Successful developments in this direction were made 
by utilizing a pre-trained wav2vec 2 ~\cite{baevski2020wav2vec} as an audio encoder, and the decoder of mBart~\cite{liu2020multilingual} as a text decoder in  state-of-the-art  (SOTA) ST systems~\cite{gallego2021end, li2020multilingual}. 

However,~this method suffers from two inherent bottlenecks mainly due to the modality differences between speech and text:~(i)~Audio signals are several orders of magnitude longer than their transcripts, and they contain lots of redundancy~\cite{zhang2020adaptive}, aggravating the alignment difficulty between speech outputs and text input, and~(ii) compared to text, audio signals exhibit a much higher degree of variations caused by speaker and noise which amounts to learning difficulties. We frame both issues under~\emph{modality gap} between speech and text representations.

There are two lines of research to bridge the modality gap.
On the one hand, prior works propose feature selection modules to compress speech length. \cite{gallego2021end, wang2020fairseq} use convolutional neural networks (CNNs) to collapse a fixed number of adjacent feature vectors into a single one. Yet, CNNs only model local information, thus exposing the risk of information loss \cite{gulati2020conformer}. \cite{zhang2020adaptive, salesky2020phone,gaido2021ctc} propose to select important features dynamically, but they require transcriptions or phonemes, which may not exist for all language pairs \cite{chen2021specrec}. 
Alternatively, a text encoder can be attached to an audio encoder in a serial or parallel fashion \cite{xu2021stacked, liu2020bridging, wang2020bridging} or be integrated with an acoustic encoder, forming a unified encoder \cite{ye2021end, tang2021improving, zheng2021fused,han2021learning}. However, this approach is either coupled with a feature selection module, or requires large amounts of ASR and MT data, which makes training inefficient.  

\begin{figure}[t]
  \centering
    \includegraphics[width=0.7\linewidth]{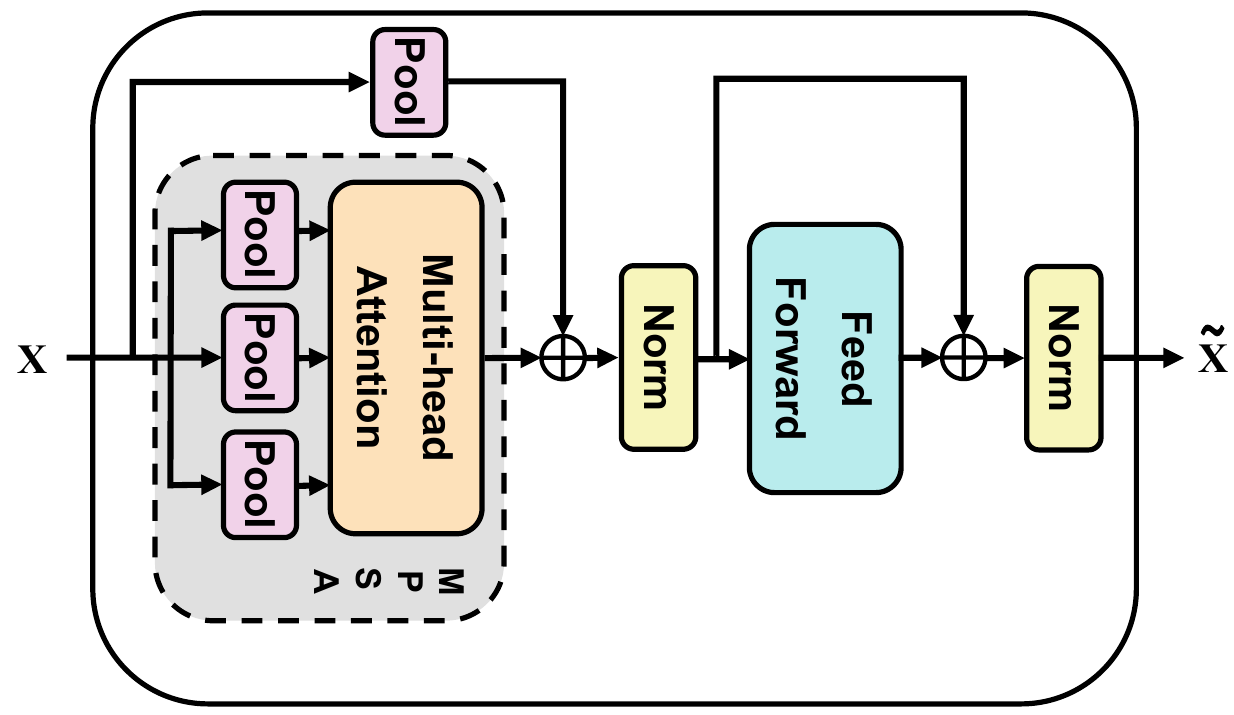}
  \caption{Overview of M-Adapter.}
  \label{fig:arch}
\end{figure}


In this work, we aim to lift the modality gap between speech and text via reducing speech length and generating features more desirable for translation. Considering that Transformer \cite{vaswani2017attention} has achieved huge success in applications in various domains, we are motivated to unleash its potential further by adapting it to be a modality adapter.
We propose M-Adapter, a Transformer-based architecture combined with convolutional layers, which models global and local dependencies of speech features while shrinking speech sequences. Figure \ref{fig:arch} depicts M-Adapter. 
The key to M-Adapter is Multi-head Pooled Self-Attention (MPSA), inside which convolution layers pool $Q$, $K$, $V$ matrices, linearly projected from input $X$, to intermediate matrices. An additional pooling module is applied to $X$. Together, they reduce the dimentionality of $X$  
as the output of the current layer, mitigating the aformentioned length mismatch issue. 
Moreover, Transformer exploits long-range global context, while convolution layer is good at modeling local features; our hypothesis is a combination of the two establishes global and local interactions, which is essential for producing higher-level features desired for ST. 

Our main contributions are: 
\begin{itemize}
    \item This is the first work on utilizing a Transformer architecture to reduce length and adapt features of a speech sequence for ST. We highlight the importance of establishing global and local interactions via self-attention and convolutions, hoping to improve our understanding of feature generation for ST.
    \item Without using any additional data, our model outperforms strong baselines across 3 Must-C language pairs, with an average improvement of 0.78 BLEU scores. 
    \item  M-Adapter helps to reduce the performance gap in various resource conditions by a large margin. 
\end{itemize}


\section{Preliminaries}
\label{sec:Preliminaries}

\subsection{Transformer}
Transformer \cite{vaswani2017attention} is a highly  modularized neural network, consisting of several Transformer blocks. Each block has two main modules, i.e., a multi-head self-attention (MSA) layer and a feedforward layer (FFN). MSA accepts an input sequence from the previous block and considers the relevance of features at other positions while it is being applied to a certain position. FFN performs non-linear transformation on these features to produce an output sequence for the next block. The two modules are wrapped by residual connection and layer normalization. Each Transformer block does not change the output sequence length. We refer readers to the original papers for details. 

\subsection{Pre-trained models}
It is not always possible to train a ST model from scratch successfully, due to limited ST corpora and computation resources. Leveraging pre-trained models trained on massive data is a promising direction, as they serve as good initiation points. It is particularly beneficial for low-resource ST. Naturally, this requires a pre-trained acoustic encoder and a pre-trained text decoder. 

\textbf{Pre-trained speech encoders.} Pre-trained models have been explored in the speech domain, to encode general-purpose knowledge. Typical methods include generative learning \cite{chorowski2019unsupervised,chen2019audio,ling2020decoar}, multi-task learning \cite{ravanelli2020multi} and discriminative learning \cite{baevski2020wav2vec,oord2018representation}. Theses acoustic models (most are Transformer-based) are trained with well-designed objective functions during the pre-training phase and used as a standalone acoustic encoder for downstream tasks. 

\textbf{Pre-trained text decoders.} Pre-training a text decoder can be done independently (e.g., GPT2 \cite{radford2019language})  or jointly with an encoder for sequence-to-sequence tasks (e.g., mT5 \cite{xue2021mt5}, mBart \cite{liu2020multilingual}). With the former approach, after the pre-training phase, the text decoder needs to be fused with additional encoders or adapters \cite{sun2021multilingual}, which increases the complexity of architectures. For the latter approach, the decoder component can usually be  used as an individual module without architectural modifications.





\section{E2E ST with M-Adapter}
\label{sec:method}

Training a cascaded system with pre-trained models is a sub-optimal option. Not only do the common problems with cascaded approaches remain unsolved, but also it requires intermediate transcripts, which is, again, not always available. Therefore, in this work we investigate how to effectively leverage pre-trained models by addressing the modality mismatch between speech and text. More specifically, we will use wav2vec 2 (W2V2) as an speech encoder and mBart decoder as the text decoder, and integrate them with our proposed M-Adapter.

\subsection{Pre-trained Modules}

\textit{W2V2} is pre-trained on large untranscribed data and it can achieve excellent transcription performance through fine-tuning on a small amount of parallel ASR data. It is pre-trained with a contrastive objective that empowers the model to distinguish a true masked segment and those produced by the model. Its Transformer layers allow the model to encode contextual information surrounding the masked segment. After the training is complete, only its CNN feature extractor and Transformer layers are kept for downstream tasks.

\textit{mBart} is a Transformer-based Seq2Seq denoising auto-encoder model, trained on massive amounts of monolingual and multilingual data. The training objective is to reconstruct an input sequence conditioned on a corrupted version. For our task, we use its decoder component.



\subsection{M-Adapter}

\begin{figure}[t]
  \centering
  \includegraphics[width=0.7\linewidth]{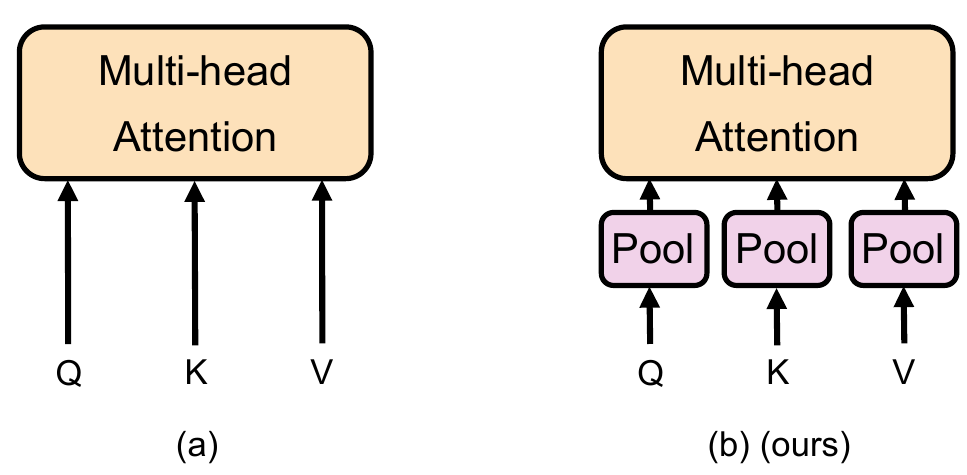}
  \caption{Multi-head Self-Attention (left) vs Multi-head Pooled Self-Attention (right)}
  \label{fig:attn}
\end{figure}

To adapt speech to text, we propose to replace MSA with MPSA, shown in Figure \ref{fig:attn}. In contrast to MSA which does not modify the sequence length, MPSA pools the sequence to reduce its length. M-Adapter aggregates local fine-grained features via pooling operation while capturing global information via MSA. 
It also encourages the model to  capture better features for the task at hand via the interactions between MSA and convolutional layers.

\textbf{Multi-head Pooled Self-Attention} Formally, given an input sequence 
$\mathbf{X} \in \mathcal{R}^{L \times D}$ where  $L$ is the sequence length and $D$ is the embedding dimension, same as MSA, MPSA first linearly projects $\mathbf{X}$ to query, key, and value matrices: $\mathbf{Q} \in \mathcal{R}^{L \times D}, \mathbf{K} \in \mathcal{R}^{L \times D}, \mathbf{V} \in \mathcal{R}^{L \times D}$. 


 Next, $\mathbf{Q}$, $\mathbf{K}$ and $\mathbf{V}$ are pooled with modules $\operatorname{Pool_Q}$, $\operatorname{Pool_K}$ and $\operatorname{Pool_V}$. Each pooling module consists of a 1D convolutional layer parameterized with kernel size $k$, stride $s$ and padding $p$. Three new matrices are obtained:
\begin{align*}
Q' = \operatorname{Pool_Q}(Q) \hspace{0.5cm} K' = \operatorname{Pool_K}(K) \hspace{0.5cm} V' = \operatorname{Pool_V}(V)
\end{align*}
where $\mathbf{Q'} \in \mathcal{R}^{L' \times D}$, $\mathbf{K'} \in \mathcal{R}^{L' \times D}$ and $\mathbf{V'} \in \mathcal{R}^{L' \times D}$ and $L'$ is the length of the new sequence, which is determined by  
\begin{align*}
    L'=\left\lfloor\frac{L+2 p - k}{s}\right\rfloor+1
\end{align*}

Attention scores are then calculated based on the new matrices as follows. 

$$
\text { Attention }(Q, K, V)=\operatorname{Softmax}\left(Q^{\prime} K^{\prime T} / \sqrt{d}\right) V^{\prime}
$$


where $\sqrt{d}$ normalizes the inner product matrix of $\mathbf{Q'}$ and $\mathbf{K'}$. Practically, we use $h$ attention heads, each of which performs attention operation on a non-overlapping subset of $d$ of the $D$ dimensions of $X$. This yields $h$ sequences which are then concatenated to $\mathbf{H} \in \mathcal{R}^{L' \times D}$.

\textbf{Pooled Input} We cannot add $\mathbf{X}$ and $\mathbf{H}$ directly (see Figure \ref{fig:arch}), as they now have different lengths. To overcome the issue, another 1D convolution pooling layer, $\operatorname{Pool_X}$, is used to convert $\mathbf{X}$ to $\mathbf{X'} \in \mathcal{R}^{L' \times D}$, to allow for subsequent addition and layer norm operations. 

Each M-Adapter layer thus reduces the sequence length by a factor of $s$.



\begin{table*}[!t]
\centering
\small
  \scalebox{0.9}
  {
  \begin{tabular}{llccccc}
    \toprule
      && Model & Encoder Size (M)  & Compression & Extra Data & BLEU \\
    \midrule
    \addlinespace[0.3em]
    \multirow{6}{4em}{\rotatebox[origin=c]{90}{\begin{tabular}{@{}c@{}}Decoder \\ Fine-tuned\end{tabular}}} & \multirow{4}{4em}{\small{Baselines}} &  W2V2-cnn-mBart$^\diamond$                                           &  117   & 8:1 & -  &26.12 \\
    & & W2V2-tran-mBart$^\diamond$                                      & 147  & 1:1 & - & 26.56  \\
    & & XSTNet$^*$                                              & 117   & 4:1 &  \checkmark &25.5 \\
    & & JT-S-MT$^+$                                           &  76 & 1:1 & \checkmark  &  26.8 \\\cline{2-7}
\addlinespace[0.3em]
  &\multirow{2}{4em}{Ours} & W2V2-mAda$_\textit{1}$-mBart & 154  & 8:1 & - & 27.00 \\ 
  & & W2V2-mAda$_\textit{3}$-mBart  & 185   &  8:1 &   & \textbf{27.13} \\
    \hhline{=======}
    \multirow{4}{4em}{\rotatebox[origin=c]{90}{\begin{tabular}{@{}c@{}}Decoder \\ Frozen\end{tabular}}} & \multirow{2}{4em}{\small{Baselines}} &  W2V2-cnn-mBart$^\diamond$                                              &  319 &  8:1 &  - & 26.45 \\
    & & W2V2-tran-mBart$^\diamond$                                            & 349  & 1:1 & - & 26.91   \\\cline{2-7}
\addlinespace[0.3em]
  & \multirow{2}{4em}{Ours} & W2V2-mAda$_\textit{1}$-mBart & 357   & 8:1 &- &27.60  \\ 
  & & W2V2-mAda$_\textit{3}$-mBart  & 386   &8:1 & - & \textbf{27.73}  \\
    \bottomrule
  \end{tabular}}
  \caption{BLEU results on Must-C EN$\rightarrow$DE.  Top rows: a decoder is fully (XSTNet and JT-S-MT) or partially (the rest) trained, while an encoder is being trained. Bottom rows: a decoder is frozen and W2V2 is fully trained. \textbf{Bold}: best results in each setting. $^\diamond$: replicated. $^*$, $^+$: taken from \cite{ye2021end}, \cite{tang2021improving}. 
  }
  \vspace{-5.5mm}
  \label{tab:focus} 
\end{table*}

\begin{table}[t]
\centering
\small
  \scalebox{0.8}
  {
  \begin{tabular}{lcccc}
    \toprule
      &  &BLEU  \\
      Model & DE & RO & FR & $\triangle$ \\
    \midrule
    \textit{Baseline}  \\
    W2V2-cnn-mBart & 26.12 & 23.72 & 36.92 &  - \\
    \midrule
   \textit{Proposed}\\
    W2V2-mAda$_\textit{3}$-mBart  & 27.13 & 24.62  & 37.34 & 0.78  \\
    \bottomrule
  \end{tabular}
  }
  \caption{
  BLEU for EN$\rightarrow$DE, RO, and FR. 
  }
  \vspace{-7mm}
  \label{tab:3lang} 
\end{table}

\section{Experiments}
\label{sec:Experiments}
\vspace{-1mm}
\subsection{Data} We experiment with Must-C \cite{cattoni2021must}, a multilingual ST corpus built from Ted talks whose source language is English (EN). We focus on EN$\rightarrow$DE~(German), RO~(Romanian ) and FR~(French). We follow~the instructions in~\cite{gallego2021end}~to preprocess data.~We develop our models on the dev set and report results on the tst-COMMON set. No transcripts or extra ASR/MT data are used.

\vspace{-1mm}
\subsection{Experimental Settings}
\subsubsection{Implementation details}
W2V2 models are pre-trained on large unlableled speech data and can be optionally fine-tuned on ASR data before used for downstream tasks. The model we use has a ``large'' architecture with the quantization module removed, pre-trained on 53.2k-hour of unlabelled data and fine-tuned on 960-hour of labelled data and pseudo-labels. The mBart model\footnote{For the rest of our paper, we refer mBart's decoder as mBart.} we use has 12-layer Transformer blocks for the decoder, fine-tuned with the multilingual MT task (English$\rightarrow$50 languages). Similar with \cite{gallego2021end}, we employ a lightweight adapter (lit-adapter)\footnote{Its main components are two linear layers. We will drop the mention of this adapter since it is used in all experiments, including the baselines we replicated.} before M-Adapter to adapt pre-trained models to new tasks. We deploy $N$ number of M-Adapter blocks. The M-Adapter modules lead to reduced length by stride factors of $s^N$. We implement two variants of our model: i) W2V2-mAda$_\textit{1}$-mBart: it has 1 M-Adapter layer (short for mAda$_\textit{1}$) with 8-stride 1D convolutional pooling modules and kernel size and padding of 8 and 4. ii) W2V2-mAda$_\textit{3}$-mBart: this model has 3 M-Adapter layers (mAda$_\textit{3}$) with kernel size, stride, padding as 3, 2 and 1. Both models shrink an input sequence by 8. We use BLEU\footnote{\url{https://github.com/mjpost/sacreBLEU}} to measure translation quality. 

\vspace{-1mm}
\subsubsection{Training strategy}
To train our model, we follow the two-step LayerNorm and Attention training strategy (LNA-2step) used in \cite{li2020multilingual}. In the first step, we train our M-Adapter layers, together with lit-adapter, and keep W2V2 and mBart frozen. In the second step, we fine-tune the adapters and a portion of W2V2 and mBart, including layer normalization, encoder self-attention and encoder-decoder attention. 
To further quantify the benefits that our approach bring to W2V2, we experiment with training the entire W2V2 while keeping the mBart frozen. Our models are trained for 32 epochs until early stopping is reached.

\vspace{-1mm}
\subsection{Baselines} 

We compare our model with the following baselines:
\setlist[itemize]{align=parleft,left=0pt..1em}
\begin{itemize}
    \item \textbf{W2V2-cnn-mBart} \cite{gallego2021end}: It is an E2E ST models using pre-trained models with a similar architecture as ours, except that it uses 3 CNN layers as the length adapter. The same W2V2 and mBart checkpoints are employed. It has the same level of length compression as our model.
    \item \textbf{W2V2-tran-mBart}: M-Adapter layers are replaced with 3 Transformer (thus short for ``tran'') layers. 
    \item \textbf{XSTNet} \cite{ye2021end}: The encoder of the model consists of a wav2vec 2 (base) and convolution layers and one unified Transformer encoder to jointly encode speech and transcript. 
    \item \textbf{JT-S-MT} \cite{tang2021improving}: It has a unified encoder (initialized with a pre-trained speech encoder and a text encoder) to perform ST and MT tasks. It leverages massive labelled MT data.
\end{itemize}
\vspace{-2mm}

\subsection{Main Results}
Table \ref{tab:focus} shows  the results on Must-C EN$\rightarrow$DE in two conditions: decoder fined-tuned and decoder frozen. In the top rows where the decoder is fine-tuned, our models surpass the baselines significantly.~Particularly, our models surpass W2V2-cnn-mBart, the previous SOTA model on this dataset, by a large margin.\footnote{It may appear our performance gain comes from more parameters, but we show that it is not the case in Subsection \ref{sec:len}.} In the bottom rows where the decoder is fixed, while our models still outperform the baselines, our models improve BLEU scores compared to our own models from the top rows by an average of 0.60.
This indicates the effectiveness of our method in unleashing the potential of W2V2.
It also highlights the importance of the quality of speech representation for ST.
Interestingly, with deeper layers, W2V2-mAd$_\textit{3}$-mBart brings a marginal improvement compared to W2V2-mAd$_\textit{1}$-mBart in both settings.
We speculate the reason is that W2V2 produces high-level features \cite{pasad2021layer} and a single M-Adapter layer is sufficient to transform these features to more complex features desired for translation. 

We also experiment on Must-C EN$\rightarrow$RO, FR, with  3 M-Adapter layers attached and the decoder fine-tuned. Table \ref{tab:3lang} summarizes the results.\footnote{We expect to see the same trend as in Table 1 for EN $\rightarrow$ RO, FR.} On average, our method improves translation quality by 0.78 in BLEU compared to the baseline.

\section{Analysis}
\label{sec:Analysis}



\subsection{Impact of Length Reduction}\label{sec:len}
Table \ref{tab:comp} demonstrates the impact of the amount of length compression on the final results. To control our settings (e.g., model size), we experiment with W2V2-mAda$_\textit{1}$-mBart, keep the kernel size constant, and vary stride from 2 to 8. We set the compression cap to 8, because beyond 8 the length of a shrunk sequence will be less than that of the corresponding text sequence.~Given the module size, constant improvements in BLEU are observed as stride increases, implying a great deal of redundancy in speech representations and the necessity of compressing them.


\vspace{-1mm}
\begin{table}[h]
\centering
\small
  \scalebox{0.8}{
    \begin{tabular}{lcccc}
    \toprule
    Kernel & 8 & 8 & 8 & 8  \\
    \midrule
    Stride & 8 & 6 & 4 & 2 \\
    \midrule
    BLEU & 27.00 & 26.79 & 26.51 & 26.10 \\
    \bottomrule
    \end{tabular}}
    \caption{Varied degree of compression for W2V2-mAd$_\textit{1}$-mBart.
    }
    \vspace{-4mm}
    \label{tab:comp}
\end{table}

\subsection{Global and Local Interactions}
We investigate why our method is superior than the baselines.
We hypothesize that M-Adapter does more than length reduction; more fundamentally, it builds good global and local interactions, via a combination of Transformer and convolutional layers in that the former excels at modelling global context and the latter at capturing fine-grained, local features. 

\noindent\textbf{Positions of Pooling Modules}: To validate our hypothesis, we conduct ablation study on W2V2-mAda$_\textit{3}$-mBart by changing the positions of the pooling modules inside M-Adapter, i.e., pooling takes place (1) before MAS (b4p); (2) before feedforward  (b4f); and 3) after the second layernorm (b4o). 
The BLEU scores for these variants are 26.27 (b4p), 25.38 (b4f) and 24.94 (b4o), compared to the current result (27.13).
This suggests that local and global information are well blended at this position.  

\noindent\textbf{Removal of Global Capacity}: 
To further investigate the global capacity of M-Adapter, we remove MPSA at the inference time and let M-Adapter perform local aggregation only. Our results show that the performance of W2V2-mAda$_\textit{3}$-mBart drops from 27.13 to 26.84, highlighting the significance of considering long-range information in producing better speech features. MPSA, complemented by FFN and other components, performs better than bare CNN layers as in the baseline.

\noindent\textbf{Perturbation on Representations}:
For the purpose of examining the quality of representations, we deliberately perturb speech features at the inference time at two levels, i.e., the output of W2V2 and the outputs of the adapters, at ratios of 10\%, 20\% and 50\%. Table \ref{tab:robust} compares the perturbation effects on W2V2-cnn-mBart and W2V2-mAd$_\textit{3}$-mBart. On the one hand, perturbation at the W2V2-level has bigger impacts on the latter than the former. We conjecture that the reason is that perturbing vectors at this level breaks the global and local linkage established by M-Adapter. On the other hand, perturbation at the adapter-level has less influence on W2V2-mAd$_\textit{3}$-mBart than W2V2-cnn-mBart. We believe this is because each vector produced by M-Adapter contains both local (with convolution layer) and global (with self-attention) information. When vectors are corrupted, the information carried in them can be inferred from the uncorrupted ones.
Effectively, more information is preserved in remaining vectors with M-Adapter than CNN, thus making M-Adapter more robust.


\begin{table}[!t]
\centering
\small
  \scalebox{0.8}
  {
  \begin{tabular}{lccc}
    \toprule
      & & \multicolumn{2}{c}{Model}  \\
    Perturb.   & \% & W2V2-cnn-mBart  & W2V2-mAd$_\textit{3}$-mBart   \\
    \midrule
    none & - & 26.12 & 27.13 \\
    \midrule
    
    W2V2 
    & \begin{tabular}{@{}c@{}} 10 \\ 20 \\ 50\end{tabular} 
    & \begin{tabular}{@{}c@{}}   25.53 \textcolor{blue}{(2\%)}\\ 24.27 \textcolor{blue}{(7\%)}\\ 12.72 \textcolor{blue}{(51\%)}\\  \end{tabular} 
    & \begin{tabular}{@{}c@{}}   26.46  \textcolor{blue}{(2\%)} \\ 24.89 \textcolor{blue}{(8\%)}\\ 9.69 \textcolor{blue}{(64\%)} \end{tabular}  \\
    \midrule
    adapter
    & \begin{tabular}{@{}c@{}}   10 \\ 20 \\ 50\end{tabular} 
    & \begin{tabular}{@{}c@{}}  23.78 \textcolor{blue}{(9\%)}\\ 20.08 \textcolor{blue}{(23\%)}\\ 9.58\textcolor{blue}{(63\%)}\\  \end{tabular} 
    & \begin{tabular}{@{}c@{}}   25.25 \textcolor{blue}{(7\%)} \\ 23.00 \textcolor{blue}{(15\%)}\\ 12.71 \textcolor{blue}{(53\%)} \end{tabular}  \\
    \bottomrule 
  \end{tabular}
  }
  \caption{BLEU scores with perturbation on representations at W2V2 and Adapter outputs, with different ratios. 
    \textcolor{blue}{$(\downarrow)$} indicates performance drop compared to the original models.
  }
  \vspace{-4mm}
  \label{tab:robust} 
\end{table}



\vspace{-1mm}
\subsection{Low-Resource Settings}
\begin{figure}[!h]
\centering

\begin{tikzpicture}[thick,scale=0.7, every node/.style={scale=1}] 

\begin{axis}[
    title style={at={(0.5,0.97)},anchor=north, yshift=-0.1, draw=gray},
    height=5cm, 
    width=10cm,
    ymajorgrids, tick align=inside,
    major grid style=dashed,
    enlarge y limits={value=.1,upper},
    ymin=15, ymax=29,
    enlarge x limits=0.15,
    ylabel={BLEU},
    xlabel={Hours},
    ylabel near ticks,
    symbolic x coords={
       408, 204, 82, 41},
    xtick=data,
    ]
\legend{W2V2-cnn-mBart, W2V2-mAda$_3$-mBart}
\addplot[mark=*,blue] table[x=method,y=W2V2-cnn-mBart]{\interngram};
\addplot[mark=triangle,cyan] table[x=method,y=W2V2-mAda$_3$-mBart]{\interngram};

\end{axis}
\end{tikzpicture}
\caption{Model performance in various resource conditions.}
\label{fig:resource}
\end{figure}
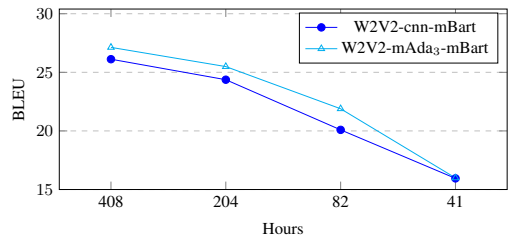

We examine the robustness of our model in different resource conditions. 
We conjecture that compressed, better quality speech representations bring more data efficiency.
We stimulate high- (408 hours), mid- (204, 82 hours) and low-resources (41 hours) settings. Figure~\ref{fig:resource} shows that in high-, mid-resource conditions, W2V2-mAd$_\textit{3}$-mBart has less degradation in BLEU, compared to W2V2-cnn-mBart, which has validated our hypothesis.
In the low-resource setting, the two models perform similarly, which is expected, as M-Adapter does require sufficient amounts of training data.
\vspace{-1mm}

\pgfplotstableread[row sep=\\,col sep=&]{
    method & W2V2-cnn-mBart & W2V2-mAda$_3$-mBart\\
408 & 26.12 & 27.13 \\
204 & 24.37 &  25.49 \\
82 & 20.09 & 21.89 \\
41 & 15.95 & 15.99 \\
    }\interngram

\vspace{-4mm}
\subsection{Sparsity}
We measure the sparsity of representations produced by above two models, with Hoyer metric \cite{hurley2009comparing}. In essence, Hoyer is a proportion of the $L^2$ over $L^1$ norm; higher Hoyer score indicates more sparsity, less utilization of the representation space. The average Hoyer scores for the two models are 0.7930 and 0.4489, respectively. This indicates that the representations produced by our model make better use of the space. 
We speculate this might serve as another indicator measuring an acoustic encoder. We will investigate further on its implications in our future work. 


    

\section{Conclusion}
\label{sec:Conclusions}
We propose M-Adapter, a novel Transformer-based architecture to lift the modality gap between speech and text representations.
We demonstrate the importance of reducing speech length and that of establishing global and local interactions via a combination of Transformer and convolution layers in producing high-level speech features that are more suitable for ST. Our method surpasses strong baselines by a large margin.


\clearpage
\bibliographystyle{IEEEtran}

\bibliography{mybib}


\end{document}